\documentclass{elsarticle}
\usepackage{stfloats}

\usepackage{times}
\usepackage{epsfig}
\usepackage{graphicx}
\usepackage{amsmath}
\usepackage{amssymb}
\usepackage{subcaption}
\usepackage{tabularx}
\usepackage{booktabs}
\usepackage{bm}
\usepackage{graphicx}
\usepackage{epstopdf}
\usepackage{color}

\usepackage{hyperref}

\begin{document}
\captionsetup[figure]{labelfont={bf},labelformat={default},labelsep=period,name={Fig.}}
\begin{frontmatter}

\title{D-VPnet: A Network for Real-time Dominant Vanishing Point Detection in Natural Scenes}
\cortext[cor]{Corresponding author}

\author{Yin-Bo Liu}

\author{Ming Zeng\corref{cor}}\ead{zengming@tju.edu.cn}
\author{Qing-Hao Meng\corref{cor}}\ead{qh\_meng@tju.edu.cn}
\address{Institute of Robotics and Autonomous Systems \\ Tianjin Key Laboratory of Process Measurement and Control \\ School of Electrical and Information Engineering\\ Tianjin University,  Tianjin 300072, China}

\begin{abstract}

As an important part of linear perspective, vanishing points (VPs) provide useful clues for mapping objects from 2D photos to 3D space. Existing methods are mainly focused on extracting structural features such as lines or contours and then clustering these features to detect VPs. However, these techniques suffer from ambiguous information due to the large number of line segments and contours detected in outdoor environments.  In this paper, we present a new convolutional neural network (CNN) to detect dominant VPs in natural scenes, i.e., the Dominant Vanishing Point detection Network (D-VPnet). The key component of our method is the feature line-segment proposal unit (FLPU), which can be directly utilized to predict the location of the dominant VP.  Moreover,  the model  also uses the two main parallel lines as an assistant to determine the position of the dominant VP. The proposed method was tested using a public dataset and a Parallel Line based Vanishing Point (PLVP) dataset. The experimental results suggest that the detection accuracy of our approach outperforms those of state-of-the-art methods under various conditions in real-time, achieving rates of 115fps. 

\end{abstract}

\end{frontmatter}

\section{Introduction}
 The vanishing point (VP) is the intersection of two parallel lines on an image from the viewpoint of linear perspective. VP detection has attracted great attention in various research  fields, such as  camera calibration \cite{caprile1990using}, 3D reconstruction \cite{guillou2000using}, pose estimation \cite{wang2001pose}, depth estimation \cite{battiato2004depth}, lane departure warning (LDW) \cite{yoo2017robust} and simultaneous localization and mapping (SLAM) \cite{ji2015rgb}. Also, it can be used to enhance the composition of photographers’ work through image re-targeting \cite{bhattacharya2010framework}, as well as provide feedback to photographers during the photographic creation \cite{yao2012oscar}.

Traditional VP detection methods  mainly focus on man-made or indoor environments , where the line segments detected from the images are either orthogonal or parallel to the direction of  gravity.  In these techniques, a large number of lines or contours are firstly detected, and then the constructed features are clustered to estimate the VPs \cite{Lezama_2014_CVPR}. However, when applied to  more complex natural scenes,  traditional methods may suffer from  ambiguous information generated by irregular line directions. Besides, the determination of the dominant VP in natural scenes is also a difficult problem.

\begin{figure}[t]
\centering
\includegraphics[scale=0.7]{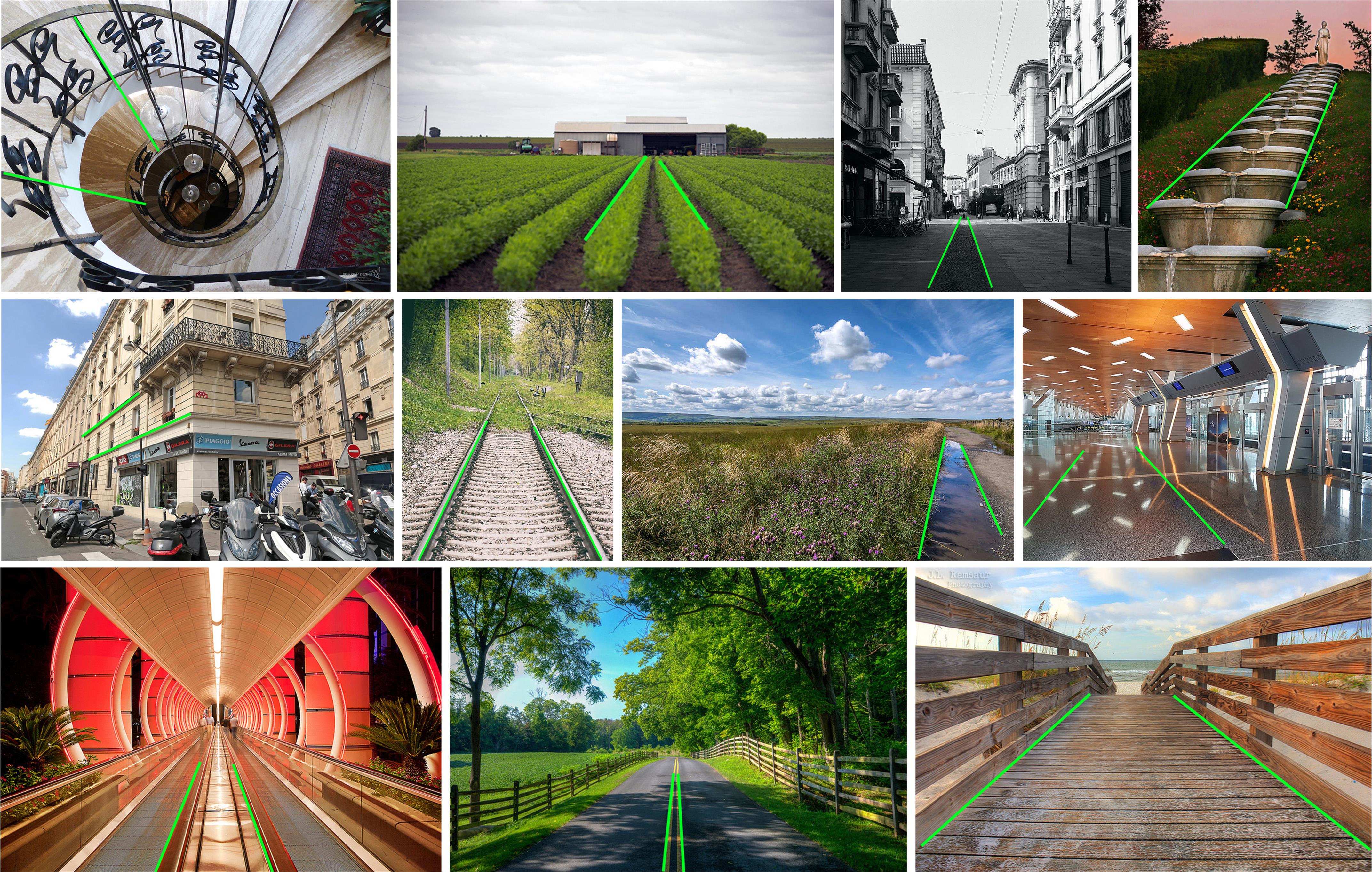}
\caption{ Dominant VPs in natural scenes. Test images from our PLVP dataset. Manually labeled ground truth lines are marked in green. }
\label{fig:label1}
\end{figure}

Specifically, a dominant VP is defined if it (i) is associated with the major geometric structures of the scene, and (ii) conveys a strong impression of 3D space or depth to the viewers \cite{zhoutmm17}. As illustrated in Fig. \ref{fig:label1}, the location of dominant VP provides important clues about the overall composition of the scene. These are utilized by photographers to optimize artistic creation and  can be applied to lane detection in Advanced Driver Assistance Systems (ADAS) \cite{ozgunalp2016TITS}. Furthermore, there is a strong demand for real-time photo composition creation for many general mobile phone users. Recently, Zhou et al. proposed a novel method for dominant VP detection by combining a contour-based  edge detector with J-Linkage \cite{zhoutmm17}. Zhang et al. derived a semantic-texture fusion network to detect the dominant VP in wild scenes \cite{zhang18}. Unfortunately, the above methods cannot achieve real-time processing because edge detection and clustering procedure are time consuming. 

 In this paper, a real-time D-VPnet model for detecting the dominant VP in natural scenes is proposed.  In particular, we use feature line-segment proposal unit (FLPU)-based CNN to directly predict the location of the dominant VP and utilize the information of the two main parallel lines  to further improve the accuracy of dominant VP detection. The main steps are as follows. Firstly, we employ the MobileNet v2 to extract the structural features from the input image.  Secondly, a three-sized grid cell detection network is  developed to quantitatively estimate the probability of a dominant VP being present for each cell. Finally, we determine the location of the final dominant VP through the largest confidence score. In the experimental analysis, we investigate the method's effectiveness systematically on a  public dataset and our constructed PLVP dataset.
 
The main contributions are as follows:
\begin{itemize}
\item We propose a single-shot deep CNN with FLPU to locate the dominant VP in natural scenes. By  combining the dominant VP features and main parallel lines features,  the proposed method significantly outperforms the traditional methods in accuracy.
\item Our approach can process 115 fps on an RTX 2080 Ti GPU, which is several times faster than state-of-the-art methods.  
\item In order to evaluate the performance of different algorithms more accurately, we have created a manually labeled dataset including 5,776 natural scene images.
\end{itemize}

The remainder of this paper is organized as follows. We first review some relevant works in Section \ref{sec2}. In Section \ref{sec3}, the proposed algorithm is introduced in detail. Then, we evaluate the performance of the proposed algorithm in Section \ref{sec4}, and followed by the conclusions in Section \ref{sec5}.

\section{Related work}\label{sec2}

We firstly introduce some traditional algorithms about VPs detection. Secondly, in consideration of using a single-shot deep CNN architecture, some state-of-the-art one-stage object detection methods are presented.

\subsection{VPs Detection}

Existing VP detection methods can be divided into two categories: line-based algorithms and CNN-based algorithms. In traditional line-based algorithms, all the line segments  extracted from the image  are clustered into one or three main directions for VP detection. Zhang et al.  \cite{zhang2016vanishing} used the RANdom SAmple Consensus (RANSAC) to cluster edges into three orthogonal line directions and find VPs based on the Manhattan world assumption. Moon et al. \cite{moon2018vanishing} applied a harmony search (HS) algorithm to detect VPs for autonomous driving, while Lezama et al. \cite{Lezama_2014_CVPR} utilized statistical significance  to find the triplet of VPs.  Antunes et al. \cite{antunes2017unsupervised} focused on the radial distortion of images and detected VPs from a single Manhattan image.

More recently, optimal techniques were introduced into VP detection. Xu et al. \cite{xu2013minimum} weighted the contribution of each line segment pair in the cluster towards the VP estimation and proposed a closed-form VP. Yang et al. \cite{yang2018fast} firstly obtained the salient texture information of the image and then used a voting approach to determine the VP. In addition, Mirzaei et al. \cite{mirzaei2011optimal} developed an  analytical approach for  computing the globally optimal estimates of orthogonal VPs . However, all the above  line-based algorithms are focused on man-made environments, which consist of numerous line segments.  In contrast, we focus on the dominant VP detection  in natural scenes, in which the number of line segments related to the dominant VP may be very small compared to irrelevant edges.  Note that little attention has been paid to natural scene dominant VPs, which provide important clues to the scene's geometry composition. Although Zhou et al. \cite{zhoutmm17} addressed this problem by exploiting global structures in natural scenes via contour detection, their algorithm still suffers from ambiguous information due to numerous irrelevant edges and image noise.

In addition, there is an increasing interest in using Convolutional Neural Networks (CNN) for VP detection. Kluger et al. \cite{kluger2017deep} used a CNN to optimize line segment extraction, which is the key step of VP detection. Zhai et al. \cite{zhai2016detecting} exploited a CNN to propose candidate horizon lines and then scored them for VP detection.  More recently, Zhang et al. \cite{zhang18} developed  a semantic-texture fusion network to detect the dominant VP  in natural landscape images.  This network includes two branches, which can be used to extract textural features and semantic features. Furthermore, the Squeeze-and-Excitation block was adopted to boost the representational power of the network.  Similar to the line-based algorithms, these methods are time-consuming due to the large number of line-segment or contour extraction processing. Besides, numerous irrelevant lines or contours also affect the accuracy of VP detection.

\subsection{CNN-based one-stage Object Detection}

In recent years, one-stage detectors have been widely studied and used for 2D object localization. One-stage approaches for object detection can directly classify and regress the candidate anchor boxes. Single shot architectures, such as the Single Shot multibox Detector (SSD) \cite{liu2016ssd} and You Only Look Once (YOLO) \cite{redmon2016you,redmon2017yolo9000,redmon2018yolov3} spurred research interest in this field. Although YOLO v3  \cite{redmon2018yolov3} has already achieved real-time performance on GPUs, it is still hard to run on mobile devices. Recently efficient neural networks were proposed to allow the detection of objects on mobile devices. An SSD with MobileNet v2 \cite{mobilenetv2} as its backbone already achieved better accuracy than YOLO v2 but with 10 times fewer parameters. ShuffleNet v2 \cite{shufflenet} and Pelee \cite{pelee} are also among the best of such solutions. Based on the accuracy and fast speed of these methods, in \cite{kehl2017ssd} the SSD was extended to predict the object’s identity, while in \cite{tekin2018real} YOLO was extended to predict a 6D object pose, and in \cite{li2019line} the RPN was extended to detect traffic lines. Previous research shows that the backbone selection has great influence on the performance of deep-learning based object detection algorithms. VGG \cite{vgg} and ResNet \cite{resnet} are widely used as backbones for generic object detection. For specific applications, we can also try some other feature extraction networks, such as stacked hourglass network \cite{newell2016stacked}, unsupervised local deep-feature alignment (LDFA) \cite{zhang2018local} and hierarchical deep word embedding (HDWE) \cite{yu2019hierarchical}.

\section{Methodology}\label{sec3}
Before designing the dominant VP detection algorithm, we consider the question: “How do humans judge the position of the dominant VP?” The common practice of humans is to first find the two main parallel lines on the image, and then determine the intersection of the two lines to determine the position of the dominant VP. Inspired by human experience in determining the dominant VP, the algorithm also introduces the main parallel line detection link to help determine the position of the dominant VP and improve the accuracy of detection. We now describe our network architecture and explain various aspects of our approach in detail.

The main challenge of the algorithm is to design a CNN that can directly estimate the coordinates of the dominant VP. As mentioned above, the dominant VP is highly dependent on the main parallel lines. Therefore, the CNN network designed in this paper comprehensively utilizes the dominant VP estimation information and the information of main parallel lines to obtain more accurate estimation results. For specific applications, the modification of the loss function is a commonly used technique for optimization of deep learning algorithms \cite{yu2019spatial}. The earlier version of our algorithm designed a loss function only concerning about the offsets between predicted dominant VP coordinates and manually labeled ground-truth coordinates, and then used this loss function to train the dominant VP detection network. But we have found that two main parallel lines related to the dominant VP is also a very valuable clue for detecting dominant VP. Two related main parallel lines are estimated based on the information of the predicted dominant VP. The loss function was modified by adding deviation information between estimated main parallel lines and manually labeled ground-truth parallel lines. Therefore, the prediction accuracy of the dominant VP has been improved. The main difference between the proposed algorithm and other algorithms is that there is no explicit edge or contour extraction process. The algorithm only uses edge or contour information implicitly to estimate the dominant VP and main parallel lines through end-to-end learning.

\begin{figure*}[t]
\centering
\includegraphics[scale=0.7]{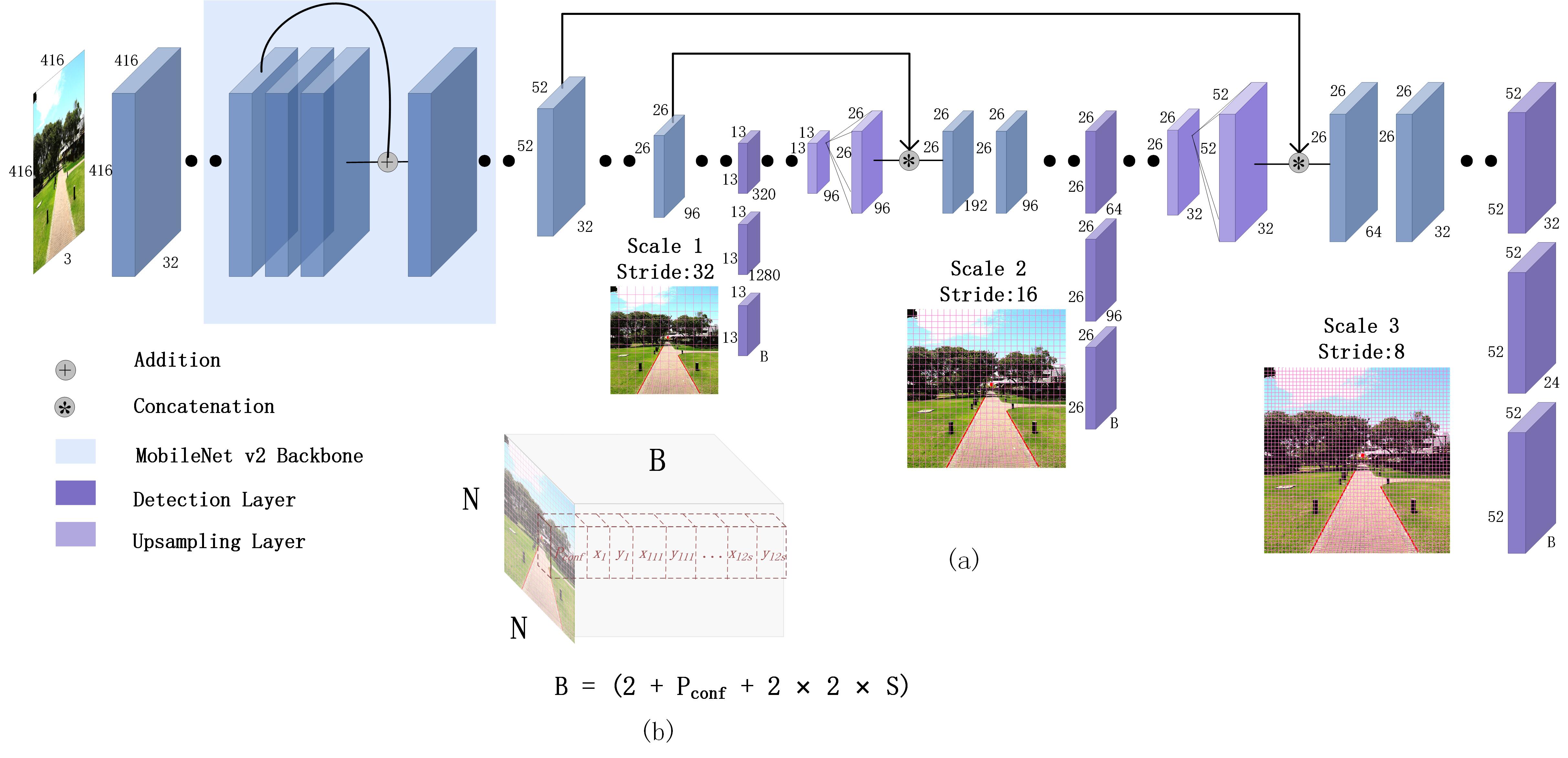}
\caption{Illustration of our D-VPnet. Our network can predict the dominant VP in 3 scales, i.e. 13 $ \times $ 13, 26 $ \times $ 26, and 52 $ \times $ 52. The 3D output tensor of our network is given in (b), and consists of the 2D location of VP and the pair of parallel lines focused on it, as well as a confidence value associated with the prediction.}
\label{fig:label2}
\end{figure*}

\subsection{Network architecture}
Inspired by single-shot object recognition networks (such as SSD and YOLO), our method divides the input image into $N \times N$ grid cells. Each grid cell predicts whether there is a dominant VP, and the prediction's values include the coordinates of the dominant VP, the predicted confidence score and the coordinates of the main parallel lines predicted by FLPU. Previous literature shows that multi-scale processing is an important technique in various image processing and computer vision applications, such as image quality assessment \cite{min2017blind,min2018blind,min2018objective}, deep learning based object detection \cite{redmon2018yolov3} and semantic segmentation \cite{zhao2019multi}. In view of the excellent performance of multi-scale processing, our network predicts VP at three different scales, which are obtained by downsampling the dimensions of the input image by 32, 16 and 8 respectively. Finally, we use the non-maximum suppression method to obtain the optimal coordinates of the dominant VP. In order to achieve a balance between accuracy and speed, our D-VPnet uses a pre-trained MobileNet v2 \cite{mobilenetv2} as the backbone, the details which are shown in Table \ref{tab1}. Overall, the output of our network at each scale is a tensor of $ N  \times N \times B$, where $B = (2 + P_{conf} + 2  \times  2  \times S)$, $S$ represents the number of points discretized by the main parallel line, and $P_{conf}$  is  the confidence score that the grid contains the dominant VP. The details are shown in Fig. \ref{fig:label2}.

\begin{table}
\small
\caption{Details of the D-VPnet. Each line describes a sequence of 1 or more identical layers, repeated $n$ times. All layers inthe same sequence have the same number $c$ of output channels. The first layer of each sequence has a stride $s$ and all others use stride 1. All spatial convolutions use $3 \times 3$ kernels. The expansion factor $t$ is always applied to the input size as described in MobileNet v2. Operator bottleneck is the module in MobileNet v2, while yolo is the block in YOLO v3.}
\begin{center}
\begin{tabularx}{7.3cm}{llllll} 
\hline
Input & Operator & t & c & n & s \\
\hline
$416^{2} \times 3 $ & conv2d     & -  & 32 & 1 & 2 \\
$208^{2} \times 32$ & bottleneck & 1  & 16 & 1 & 1 \\
$208^{2} \times 16$ & bottleneck & 6  & 24 & 2 & 2 \\
$104^{2} \times 24$ & bottleneck & 6  & 32 & 3 & 2 \\
$52^{2} \times 32$  & bottleneck & 6  & 64 & 4 & 2 \\
$26^{2} \times 64$  & bottleneck & 6  & 96 & 3 & 1 \\
$26^{2} \times 96$  & bottleneck & 6  & 160 & 3 & 2 \\
$13^{2} \times 160$ & bottleneck & 6  & 320 & 1 & 1 \\
$13^{2} \times 320$ & bottleneck & 6  & 1280 & 1 & 1 \\
$13^{2} \times 1280$ & yolo & -  & 320 & 3 & - \\
$13^{2} \times 1280$ & detection & -  & B & 1 & - \\
$26^{2} \times 96$   & yolo & -  & 64 & 3 & - \\
$26^{2} \times 96$ & detection & -  & B & 1 & - \\
$52^{2} \times 32$   & yolo & -  & 24 & 3 & - \\
$52^{2} \times 32$ & detection & -  & B & 1 & - \\
\hline
\end{tabularx}
\label{tab1}
\end{center}
\end{table}

\subsection{Feature Line-segment Proposal Unit}

As mentioned above, if all edges and contours in the image are extracted explicitly, irregular edges and contours may contain a large amount of ambiguous information, which will pose challenges in estimating the dominant VP. Therefore, the FLPU is proposed to solve the above problems. In general, there may be many parallel lines in a graph that can be used to detect the dominant VP, but it is not necessary to find all the main parallel lines, as only one representative set is needed.

In the estimation of the main parallel line segment, we use multi-discrete point prediction instead of direct line segment estimation. When the number of discrete points is sufficient, the estimated parallel line is more reliable than the method of estimating the parallel line using the starting and end points. As illustrated in Fig. \ref{fig:label2}, we divide the pair of  parallel lines into $S-1$ segments; therefore, each parallel line \bm{$l$} can be expressed as a sequence of 2D coordinates, namely \bm{${l=\{p_{sl},p_{sl+1},\cdots,p_{el}\}}$},where $p_{sl}$ denotes the coordinate of the start-index for parallel line \bm{$l$}, and $p_{el}$ represents its end-index coordinate. Similarly, a line proposal (i.e. straight line) \bm{$L$} can be written as \bm{$L=\{P_{sL},P_{sL+1},\cdots,P_{el}\}$}.

As proven in YOLO v2 and v3, prediction of the offset to the cell is a stabilized and accurate way in 2D coordinate regression, so we used the method of coordinate prediction of YOLO v3. Thus, the predicted point $(V_{x},V_{y})$ is defined as 
\begin{gather}
V_{x}=f(x)+c_{x} \\
V_{y}=f(y)+c_{y}
\end{gather}

\noindent where $f(\cdot)$  is a sigmoid function in the case of dominant VP and an identity function in the case of parallel line segment. $(c_{x},c_{y})$ is the coordinate of the top-left corner of the associated grid cell.

\subsection{Loss Function}

To train our complete network, we minimize the following loss function.
\begin{equation}
L_{vp}=\lambda_{coord}l_{coord}+\lambda_{conf}l_{conf}+\lambda_{l}(l_{ll}+l_{lr})
\end{equation}
\begin{equation}
Err(\bm{L}, \bm{l})=\frac{\sum_{x=sl}^{el} \left | P_{x}-p_{x} \right |}{el-sl+1}
\end{equation}

\noindent where $l_{coord}, l_{ll}, l_{lr}, l_{conf}$ and $Err(\bm{L}, \bm{l})$ denote the coordinate losses of the dominant VP, the left line, the right line, the confidence loss and line's coordinate error, respectively. We use the mean-squared error for the VP coordinate loss and line loss, while as suggested in YOLO v3 \cite{redmon2018yolov3}, confidence, is predicted through logistic regression. For the cells that do not contain VP, we set $ \lambda_{conf} $ to 0.5, and for the cell that contains VP we set $ \lambda_{conf} $ to 1. $ \lambda_{coord} $ and $ \lambda_{l} $ are used for balancing the training between the accuracy and coverage of the VP. Here, we set $ \lambda_{l} $ to $2.5/S$, where $S$ is the slice number of the line.

\subsection{Implementation Details}

We implemented our method in Python using Pytorch 0.4 and CUDA 10 and ran it on an i7-8700K@3.7GHz with dual NVIDIA RTX 2080 Ti. We used our PLVP training dataset, which contains 4,382 images, and both the PLVP test portion of the dataset as well as Flickr in \cite{zhoutmm17} as the test datasets. All input images were reshaped to 416 $ \times $ 416 for both training and testing. We initialized the parameters of our backbone network with a pre-trained MobileNet v2, which was pre-trained on ImageNet classification. We used stochastic gradient descent (SGD) for optimization and started with a learning rate of 0.001 for the backbone and 0.01 for the rest of the network. We divided the learning rate by 10 every 20 epochs, with a momentum of 0.9. We also used dataset augmentation with random flip and image rotations. Our code will be made publicly available for the sake of reproducibility.

\section{Experiments}\label{sec4}

In this section, we first introduce the construction of our own PLVP dataset, and then illustrate a comprehensive performance study of our proposed method and compare it to state-of-the-art algorithms using Zhou's public dataset and our own PLVP dataset. Subsequently, we quantitatively analyze the influence of each part of the model on performance through an ablation study.

\begin{figure*}[htbp]
\centering
\includegraphics[scale=1.6]{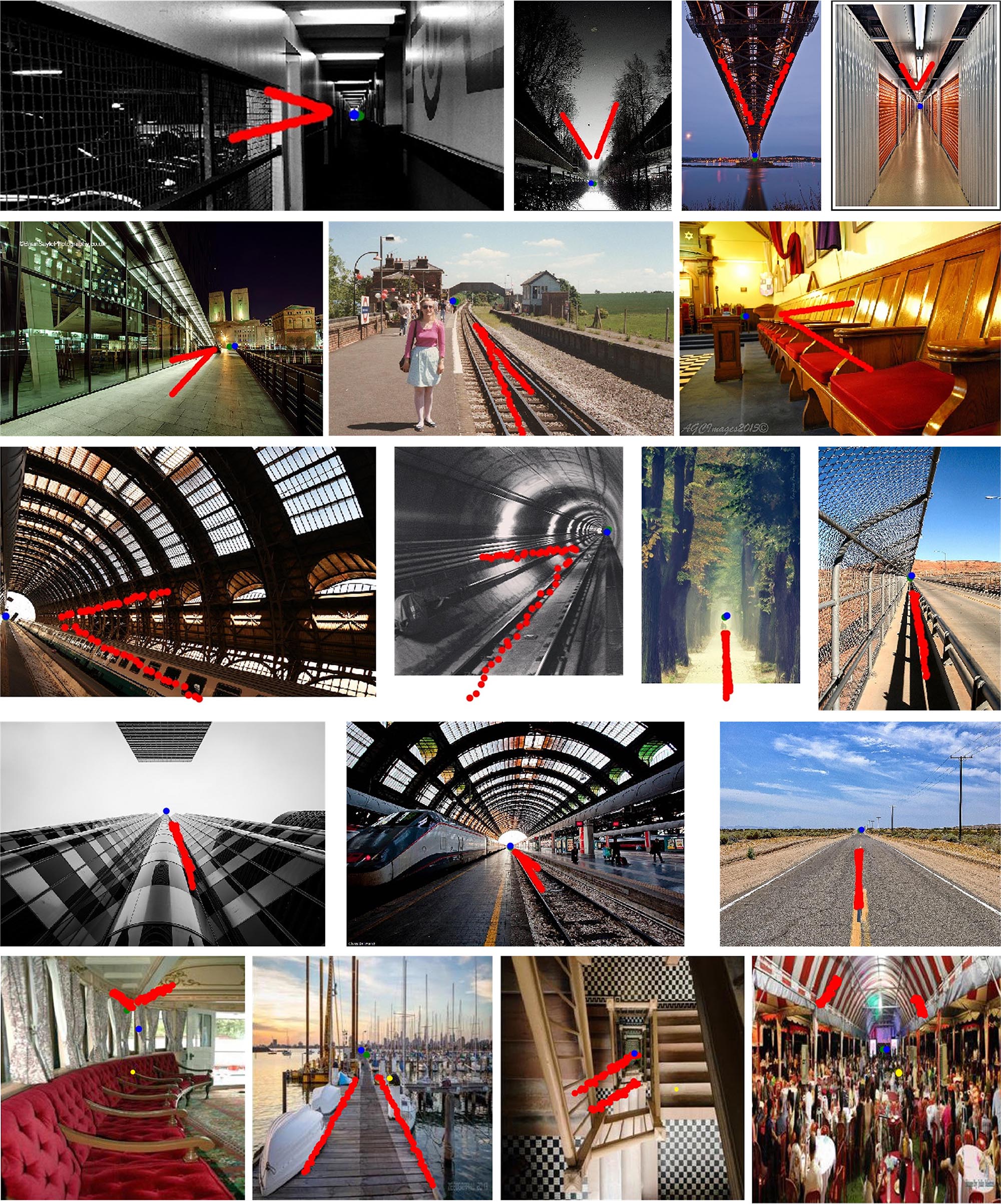}
\caption{Dominant VPs detection using our method, where the blue points are the prediction of our network, yellow points are the predictions of \cite{zhang18}, red points are the FLPU points of the predicted VPs and green points are the ground truth VPs. Images on the last row are the failure cases of Zhang's \cite{zhang18} method.}
\label{fig:label3}
\end{figure*}

\subsection{Dataset construction}
In order to evaluate the performance of dominant VP detection algorithms, we need to test the proposed approach and state-of-the-art methods on public datasets. At present, there are three kinds of public datasets, i.e., the York Urban Dataset (YUD) \cite{denis2008efficient}, the Eurasian Cities Dataset (ECD) \cite{tretyak2012geometric} and Zhou's dataset \cite{zhoutmm17}, which can be used to evaluate VP detection algorithms. We found that these public datasets have certain problems. The purpose of this paper is to evaluate the performance of the algorithm for dominant VP detection, but the YUD and ECD datasets are only suitable for evaluation of multiple VPs in urban scenes. Although Zhou's dataset can be used to evaluate  dominant VP detection algorithms, the number of images in the dataset is small, making it difficult to obtain objective evaluation results. In view of the abovementioned problems with the public datasets, we have utilized Flickr and Aesthetic Visual Analysis (AVA) \cite{murray2012ava} tools to construct a new dataset that can be used to evaluate different algorithms comprehensively and objectively.

Without using the information of two main parallel lines, the location of the dominant VP cannot be labeled accurately. Therefore, we first label the main parallel lines on the image, which are a set of parallel lines representing the main structure in the scene, denoted as \bm{$l_{1}$} and \bm{$l_{2}$}. Then, the location of the dominant VP can be calculated as \bm{$v=l_{1} \times l_{2}$}. To label the main parallel lines, we follow three principles: First, because we adopt the CNN object detection architecture in our model, and this architecture cannot detect  dominant VPs outside the images, we remove the pictures where the dominant VP lies outside the image. Second, since our goal is to establish a dataset for the evaluation of dominant VP detection, we remove images with multiple vanishing points. Finally, the labeled starting and ending points must be in the main part of the image structure. In the image content retrieval, we obtained more than 20,000 images using keywords, such as vanishing point, landscape, road, urban, from Flickr, and more than 5,000 images from the AVA dataset. Based on the above principles, we obtained our PLVP dataset, which includes  5,776 labeled images, from which 4,382 images were used as the training set and 1,394  images were used as the test set. Some samples of our dataset are shown in Fig. \ref{fig:label1}.
\subsection{Metrics}

As mentioned earlier, only Zhou's and our dataset can be used to evaluate the performance of the dominant VP algorithm. However, these two datasets partially overlap, so we only chose the Flickr part of Zhou's dataset for testing. In addition, we deleted some of the vanishing points outside of the images in Zhou's dataset, finally obtaining 840 images that met the requirements. Some dominant VP detection results are illustrated in Fig. \ref{fig:label3}.

The metric of consistency error is often utilized to evaluate the performance of VPs detectors, and is defined as follows:

\begin{gather}
err(\bm{\widehat{v}})=\dfrac{1}{K}\sum_{k}D_{RMS}(E_{k}^{G},\bm{\widehat{v}}) \\
D_{RMS}(E_{i},\bm{v_{j}})=\min \limits_{ l:l\times v_{j}=0}\left(\dfrac{1}{N}\sum_{\emph{\textbf{p}}\in E_{i}} dist(\textbf{p},\textbf{l})^2\right)^{\frac{1}{2}}
\end{gather}
where $D_{RMS}(E_{i},\bm{v_{j}})$ is a measure of consistency between edge $E_{i}$ and VP hypothesis $\bm{v_{j}}$,             
 $D_{RMS}$ is the root mean square (RMS) distance from all points on $E_{i}$  to a line $\emph{\textbf{l}}$, such that $\emph{\textbf{l}}$ passes through $\emph{\textbf{v}}$ and minimizes the distance. $N$ is the number of points on $E_{i}$, while $dist(\emph{\textbf{p}},\emph{\textbf{l}})$ is the perpendicular distance from a point $\emph{\textbf{p}}$ to a line $\emph{\textbf{l}}$. Let ${\{E_{k}^{G}\}_{k=1}^{K}}$ be the set of ground truth edges, and $err(\bm{\widehat{v}})$ be the consistency error of a detection $\bm{\widehat{v}}$.

\subsection{Comparisons}

\begin{figure*}[htbp]
    \begin{minipage}[b]{0.99\linewidth}
    \centering\includegraphics[scale=0.7]{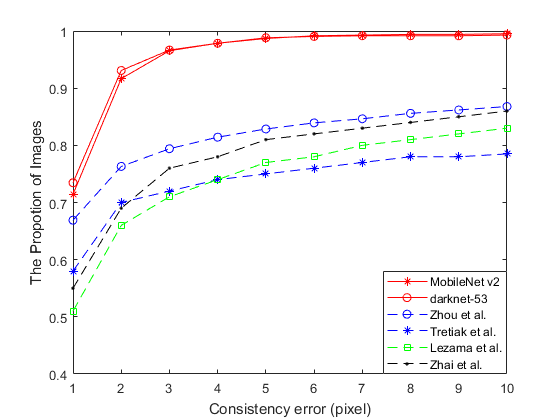}
    \subcaption{Zhou's test dataset}\label{fig4.sub.1}
    \end{minipage}
    
    \begin{minipage}[b]{.99\linewidth}
    \centering\includegraphics[scale=0.7]{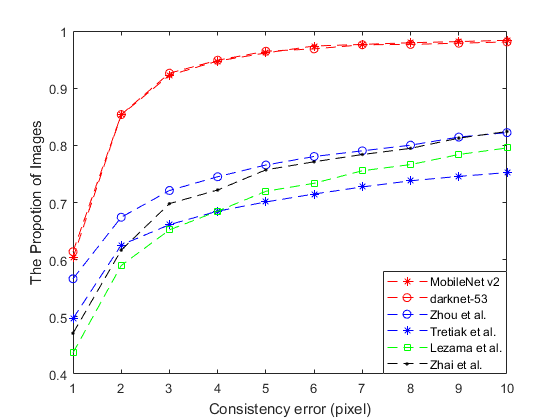}
    \subcaption{PLVP test dataset}\label{fig4.sub.2}
    \end{minipage}
    
    \caption{Experimental results of our method on Zhou's and the PLVP test datasets. Compared with other methods, our method clearly outperforms the state-of-the-art and can achieve higher accuracy with small consistency errors.}
    \label{fig4.main}
\end{figure*}

\subsubsection{Comparisons with State-of-the-art} Using Zhou's dataset, we compare the effectiveness of our method with four competitive approaches. The test results indicate that the accuracy of our method exceeds the best results obtained using Zhou's method, as shown in Fig. \ref{fig4.main}(a). In particular, with the proposed method, the consistency error of 71.43\% of the test dateset was less than or equal to 1, and the consistency error of 91.67\% of the images was less than or equal to 2. Our method achieved a consistency error of less than or equal to 5, in 98.69\% of the images. In contrast, using Zhou's method, which provided the best results, only 64\% of the images had a  consistency error of less than or equal to 1, and 77\% image had a consistency error of less than or equal to 2. The image coverage for a consistency error of less than or equal to 5 was 84\%.

The comparison of the proposed method and four state-of-the-art methods on our PLVP dataset are shown in Fig. \ref{fig4.main}(b). When the consistency error is less than or equal to 1, our algorithm has a 4\% higher image coverage than Zhou's method, which gave the best results of the other methods. For a consistency error less than or equal to 5, the image coverage of our technique was 21\% higher than that of the best Zhou's method. The results of the comparison analysis of the two databases show that the proposed method is significantly better than the existing state-of-the-art methods.

\begin{table}[!htbp]
\small
\caption{Comparisons with different backbones. To assess accuracy, we present consistency error data for errors less than or equal 1, 2 and 3 (CE$\le 1$,CE$\le 2$ and CE$\le 3$). For speed comparisons, we present GPU fps and CPU fps values.}
\begin{center}
\begin{tabularx}{12.2cm}{llllll} 
\hline
backbones & CE$\le 1$ & CE$\le 2$ & CE$\le 3$ & GPU-Fps & CPU-Runtime \\
\hline
ResNet-50     & 69.17\%  & 90.6\%  & 95.36\%  &20.3fps & 300ms\\
DarkNet-53    & \textbf{73.45\%}   & \textbf{93.1\%} & \textbf{96.67\%}  &\textbf{79.8fps} & \textbf{229ms}\\
\hline
MobileNet v1  & 43.81\%  & 73.93\% & 89.05\%  &106.3fps & 119ms\\
Pelee         & 63.6\%   & 88.4\%  & 93.2\%   &110.6fps & 80.1ms\\
Sharp-Pelee   & 71.5\%   & 91.7\%  & 95.9\%   &102.4fps & 82.5ms\\
MobileNet v2  & \textbf{71.43\%}  & \textbf{91.67\%} & \textbf{96.55\% } &\textbf{115.2fps} & \textbf{79.2ms}\\
\hline
\end{tabularx}
\label{tab2}
\end{center}
\end{table}

\subsubsection{Comparisons with different backbones} We experimented using different backbones to increase the precision and efficiency of our model. As can be seen from Table \ref{tab2}, when the consistency error is less than or equal to 1, the model accuracy using darknet-53 as the backbone is the highest, but only 2.02\% higher than the accuracy of when using MobileNet v2 as the backbone. However, the latter runs 2.5 times faster than the former. The test results in our PLVP data set show that the former is only 1.06\% higher than the latter, as shown in Fig. \ref{fig4.main}(b). When balancing accuracy and speed, we finally decided to use MobileNet v2 as the backbone of the network.

\subsection{Ablation Study}

The FLPU and the 3-scale detector are two important branches in our D-VPnet model. Therefore, we will quantitatively analyze the impact of these brancdiffhes on the performance of the detection network.

\subsubsection{FLPU} 
\begin{figure}[htbp]
\centering
\includegraphics[scale=0.80]{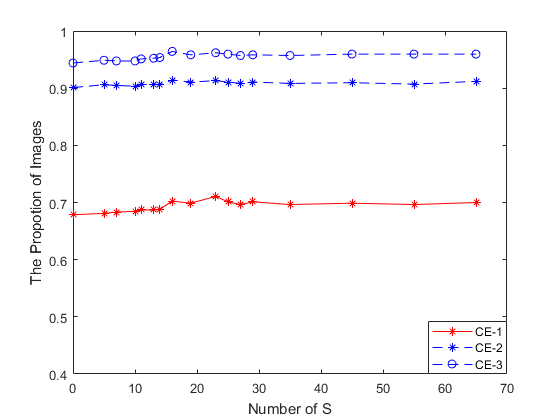}
\caption{Ablations on VP detection of FLPU. The best accuracy of the network is achieved when $S = 23$, after that the network becomes slow and accuracy stops rising.}
\label{fig:label5}
\end{figure}

\begin{figure}[htbp]
\centering
\includegraphics[scale=0.7]{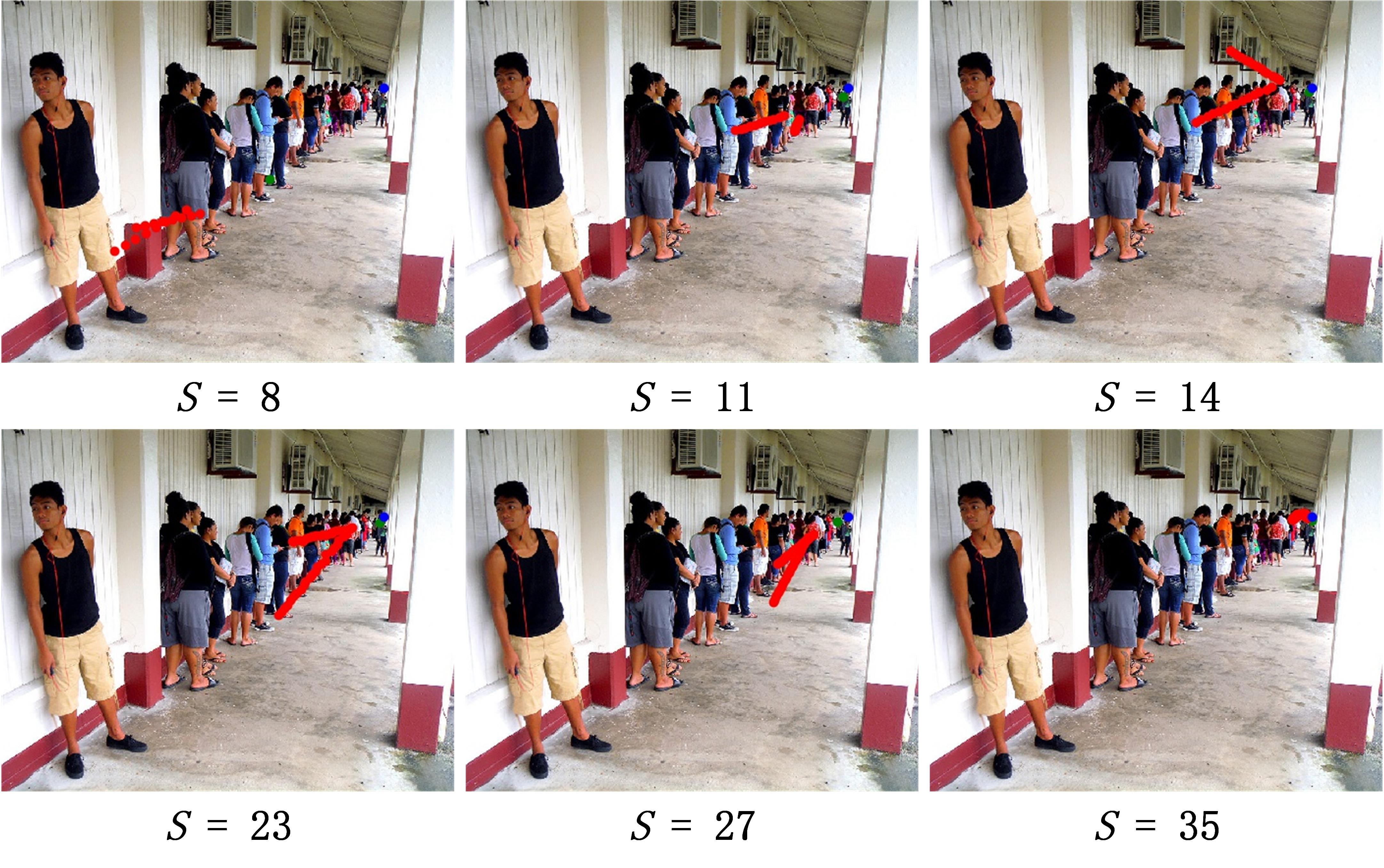}
\caption{Ablations on VP detection of FLPU. The FLPU performance is shown for $S = 8, 11, 14, 23, 27$ and $35$.}
\label{fig:label6}
\end{figure}

In the FLPU, the optimal selection of line segment number $S$ is crucial. The influence of the different values of $S$ on the performance of the model is shown in Fig. \ref{fig:label5} When the consistency error (CE)$\ge 3$, the image coverage is close to 100\%, and the contribution of the FLPU is of little significance. Therefore, we only present the results of CE$\le 1$, CE$\le 2$ and CE$\le 3$. It can be seen from Fig. \ref{fig:label5} that the accuracy gradually increases with the increase of S, but when $S$ exceeds a certain value, the accuracy no longer rises and instead falls. The effect on the actual detection is shown in Fig. \ref{fig:label6}. When $S$ is small ($S=8$), the correlation between the dominant VP predicted by the network and the main parallel lines predicted by the FLPU is unstable. When the value of $S$ is too large ($S=35$), the main parallel lines may not be detected, so corresponding information cannot be effectively utilized to assist in predicting the position of the dominant VP. Through a large number of tests, it was found that the model yields better detection results when $S=23$.

\begin{table}[!htbp]
\small
\caption{Ablations on VP detection of multi-scale detectors.}
\begin{center}
\begin{tabularx}{7.5cm}{lllll} 
\hline
Scales  & CE$\le 1$ & CE$\le 2$ & CE$\le 3$  \\

\hline
Scale-1     & 65.83\%  & 90.0\%  & 95.71\%  \\
Scale-2     & 63.5\%  & 88.3\%  & 93.4\%  \\
Scale-1,2   & 68.45\%  & 90.95\% & 95.6\%  \\
Scale-3     & 57.26\%  & 84.76\% & 92.62\%  \\
Scale-1,2,3 & \textbf{71.07\%}  & \textbf{91.30\%} & \textbf{96.19\%} \\
\hline
\end{tabularx}
\label{tab3} 
\end{center}
\end{table}

\subsubsection{3-Scale detector}In order to verify the effectiveness of the multiscale processing, we conducted several comparative experiments. Firstly, three single-scale object detection experiments were implemented, and the results showed that the model using scale-1 features was better. Subsequently, the experiments on the combination of scale-1and scale-2 were performed, and the results demonstrated that the model using two-scale features slightly outperformed the best single-scale model. Finally, three-scale combination experiments were conducted, and it was found that the detection results of the three-scale combination model were significantly better than those models with single-scale and two-scale combination. The performance of the 3-scale detector on the dominant VP detection is presented in Table \ref{tab3}.

\section{Conclusions}\label{sec5}

Dominant VP detection is a challenging problem of significant importance, especially for the photographic creation and lane detection. In this paper, we  present a D-VPnet model for dominant VP detection in natural scenes. Experimental results show that the proposed method achieves high accuracy and robustness under various conditions. In addition, our approach achieves 115 fps on an RTX 2080 Ti GPU which is several times faster than that of traditional methods.  Therefore, it is suitable for many real-time applications, such as autonomous navigation in unstructured road scenes. In the future, we plan to apply the proposed algorithm to real time pose estimation and SLAM. Moreover, we will make our PLVP dataset publicly available. Whether the proposed method can be applied to other types of images (e.g., computer graphic images \cite{min2017unified}) is an interesting topic and we plan to construct several special datasets to test the extension performance of our proposed method. Note that our model is based on an object detection CNN architecture. Therefore, one limitation of our approach is that it cannot be used to detect dominant VPs outside the images. 

\section*{Acknowledgment}
This work is supported by the National Natural Science Foundation of China (No. 61573253), and National Key R\&D Program of China under Grant No. 2017YFC0306200.

\bibliography{main}
\end{document}